\documentclass{article}

\usepackage{microtype}
\usepackage{graphicx}
\usepackage{subfigure}
\usepackage{booktabs}
\usepackage{amsfonts,xcolor}

\usepackage{hyperref}
\usepackage{mathtools}

\newcommand{\Mc}{\mathcal{M}}
\newcommand{\R}{\mathbb{R}}
\newcommand{\Lb}{\mathbf{L}}
\newcommand{\hf}{\widehat{f}}
\newcommand{\hh}{\widehat{h}}

\usepackage[accepted]{icml2022_TAGML}
\icmltitlerunning{The Manifold Scattering Transform for High-Dimensional Point Cloud Data}

\begin{document}

\twocolumn[
\icmltitle{The Manifold Scattering Transform for High-Dimensional Point Cloud Data}



\icmlsetsymbol{equal}{*}
\begin{icmlauthorlist}
\icmlauthor{Joyce Chew}{equal,U}
\icmlauthor{Holly R. Steach}{equal,YG}
\icmlauthor{Siddharth Viswanath}{UCI}
\icmlauthor{Hau-Tieng Wu}{D,D2}
\icmlauthor{Matthew Hirn}{MSU,MSU1}
\icmlauthor{Deanna Needell}{U}
\icmlauthor{Smita Krishnaswamy}{equal,YG,YCS,YAM}
\icmlauthor{Michael Perlmutter}{equal,U}
\icmlcorrespondingauthor{Michael Perlmutter}{perlmutter@ucla.edu}
\end{icmlauthorlist}

\icmlaffiliation{UCI}{UC Irvine Department of Computer Science, Irvine, CA, USA}
\icmlaffiliation{U}{UCLA Department of Mathematics, Los Angeles, CA, USA}
\icmlaffiliation{D}{Duke University, Department of Mathematics, Durham, NC, USA}
\icmlaffiliation{D2}{Duke University, Department of Statistical Science, Durham, NC, USA}
\icmlaffiliation{MSU}{Michigan State University, Department of Mathematics, East Lansing, USA}
\icmlaffiliation{MSU1}{Michigan State University, Department of CMSE, East Lansing, USA}
\icmlaffiliation{YG}{Yale University, Department of Genetics, New Haven, CT, USA}
\icmlaffiliation{YCS}{Yale University, Department of Computer Science, New Haven, CT, USA}
\icmlaffiliation{YAM}{Yale University, Applied Math Program, New Haven, CT, USA}

\icmlkeywords{Machine Learning, ICML}

\vskip 0.3in
]



\printAffiliationsAndNotice{\icmlEqualContribution} 

\begin{abstract}The manifold scattering transform is a deep  feature extractor for data defined on a Riemannian manifold. It is one of the first examples of extending convolutional neural network-like operators to general manifolds. The initial work on this model focused primarily on its theoretical stability and invariance properties but did  not provide methods for its numerical implementation except in the case of two-dimensional surfaces with predefined meshes. In this work, we present practical schemes,  based on the theory of diffusion maps, for implementing the manifold scattering transform to datasets arising in naturalistic systems, such as single cell genetics, where the data is a high-dimensional point cloud modeled as lying on a low-dimensional manifold. We show that our methods are effective for signal classification and manifold classification tasks. 

\end{abstract}

\section{Introduction}

The field of \emph{geometric deep learning},  \cite{Bronstein:geoDeepLearn2017} aims to extend the success of neural networks and related architectures to data sets with non-Euclidean structure such as graphs and manifolds. On graphs, convolution can be defined either through localized spatial aggregations or through the spectral decomposition of the graph Laplacian. In recent years, researchers have used these notions of convolutions to introduce a  plethora of neural networks for graph-structured data \citep{wu2020comprehensive}. Similarly, on manifolds, convolution can be defined via local patches \citep{Masci:geoCNN2015}, or via the spectral decomposition of the Laplace Beltrami operator \citep{boscaini2015learning}.  However, neural networks for high-dimensional manifold-structured data are much less explored, and the vast majority of the existing research on manifold neural networks 
focuses on two-dimensional surfaces.

In this paper, we are motivated by the increasing presence of high-dimensional data sets arising in biomedical applications such as single-cell transcriptomics, single-cell proteomics, patient data, or neuronal activation data \citep{tong2021diffusion,moon2018manifold,wang2020novel}. Emergent high-throughput technologies are able to measure proteins at single-cell resolution and present novel opportunities to model and characterize cellular manifolds where combinatorial protein expression defines an individual cell's unique identity and function.   
In such settings, the data is often naturally modelled as a point cloud $\{x_i\}_{i=0}^{N-1}\subseteq \mathbb{R}^n$ lying on a $d$-dimensional Riemannian manifold $\mathcal{M}$ some $d\ll n$. The purpose of this paper is to present two effective numerical methods, based on the theory of diffusion maps \citep{coifman:diffusionMaps2006}, for implementing the manifold scattering transform, a model of spectral manifold neural networks, for data living on such a point cloud.  This is the first work on the scattering transform to focus on such point clouds.

The original, Euclidean  scattering transform  was introduced in  \citet{mallat:firstScat2010,mallat:scattering2012} as a means of improving our mathematical understanding of deep convolutional networks and their learned features. Similar to a convolutional neural network (CNN), the scattering transform is based on a cascade of convolutional filters and simple pointwise nonlinearities. The principle difference between the scattering transform and other deep learning architectures is that it does not learn its filters from data. Instead, it uses predesigned wavelet filters. As shown in~ \cite{mallat:scattering2012}, the resulting network is provably invariant to the actions of certain Lie groups, such as the translation group, and is Lipschitz stable to diffeomorphisms which are close to translations. 

While the initial motivation of \citet{mallat:scattering2012} was to understand CNNs, the scattering transform can also be viewed as a multiscale, nonlinear feature extractor. In particular, for a given signal of interest, the scattering transform produces a sequence of features called scattering moments (or scattering coefficients). Therefore, the methods we will present here can be viewed as a method for extracting geometrically informative features from high-dimensional point cloud data. In addition, similar to the way GNNs can offer embeddings of nodes as well as the entire graphs, the manifold scattering transform can offer an embedding of points in the cloud as well as the entire point cloud. 

Recently, several works  have proposed versions of the scattering transform for graphs \citep{gama:diffScatGraphs2018,zou:graphCNNScat2018,gao:graphScat2018} and manifolds \citep{perlmutter:geoScatCompactManifold2020} using different wavelet constructions. 
  \citet{gao:graphScat2018} uses wavelets constructed from the lazy random walk matrix $P=\frac{1}{2}(I+WD^{-1})$, where $W$ is the weighted adjacency matrix of the graph and $D$ is the corresponding degree matrix, and \citet{gama:diffScatGraphs2018} uses wavelets constructed from $T=D^{-1/2}PD^{1/2}$. (The wavelets used in \citet{zou:graphCNNScat2018} are based off of \citet{hammond:graphWavelets2011} and are substantially different than other wavelets discussed here which are based off of \citet{coifman:diffWavelets2006}.) 
 
 Here, we present a manifold scattering transform which uses wavelets based on the heat semigroup $\{H_t\}_{t\geq 0}$ on $\mathcal{M}$. The semigroup describes the transition probabilities of a Brownian motion on $\mathcal{M}$. Therefore our wavelets have a similar probabilistic interpretation to those used in \citet{gao:graphScat2018}. 
 
In this work, we do not assume that we have global access to the manifold and our data does not come pre-equipped with a graph structure. Instead, we construct a weighted graph with weighting schemes based on \citet{coifman:diffusionMaps2006} that aim to ensure that the spectral geometry of the graph mimics the spectral geometry of the underlying manifold. We then implement discrete approximations of the heat semigroup which are not, in general, equivalent to a lazy random walk on a graph. 

Our construction improves upon the manifold scattering transform presented in \citet{perlmutter:geoScatCompactManifold2020} by using a more computationally efficient family of wavelets as well as formulating methods for embedding and classifying entire manifolds. The focus of  \citet{perlmutter:geoScatCompactManifold2020} was primarily proposing a theoretical framework for learning on manifolds and showing that the manifold scattering transform had similar stability and invariance properties to its Euclidean counterpart. Thus, the numerical experiments in \citet{perlmutter:geoScatCompactManifold2020} were somewhat limited and only focused on two-dimensional manifolds with predefined triangular meshes. 

The purpose of this paper is to overcome this limitation by presenting effective numerical schemes for implementing the manifold scattering transform on arbitrarily high-dimensional point clouds. 
It uses methods inspired by the  theory of diffusion maps \citep{coifman:diffusionMaps2006} to approximate the Laplace Beltrami operator. 

To the best of our knowledge, this work is the first to propose deep feature learning schemes for any spectral manifold network outside of two-dimensional surfaces. Therefore, our work can also be interpreted as a blueprint for how to adapt spectral networks to higher-dimensional manifolds. For example,  the theory presented in \citet{boscaini2015learning} uses the Laplace Beltrami operator on two-dimensional surfaces and could in principle be extended to general manifolds, but the numerical methods are restricted to two-dimensional surfaces in that work. One could use the techniques presented here to extend such methods to higher-dimensional settings.

{\bf Organization.}  The rest of this paper is organized as follows. In Section \ref{sec: MSdef}, we will present the definition of the manifold scattering transform, and in Section \ref{sec: implemention}, we will present our methods for implementing it from point cloud data. In Section \ref{sec: related}, we will further discuss related work concerning geometric scattering and manifold neural networks. In Section \ref{sec: results}, we present numerical results on both synthetic and real-world datasets before concluding in Section \ref{sec: conclusion}.

{\bf Notation.}
We let $\mathcal{M}$ denote a smooth, compact,  connected $d$-dimensional Riemannian manifold without boundary isometrically embedded in $\mathbb{R}^n$ for some $n\geq d$. We let  $\Lb^2 (\Mc)$  denote the set of functions $f : \Mc \rightarrow \R$ that are square integrable with respect to the Riemannian volume $dx,$ and let  $-\Delta$ denote the (negative) Laplace-Beltrami operator on $\Mc$.  We let $\varphi_k$ and $\lambda_k\geq0$, $k\geq 0$, denote the eigenfunctions and eigenvalues of $-\Delta$ with $-\Delta\varphi_k=\lambda_k\varphi_k.$ We will let $\mathcal{W}_Jf \coloneqq \{W_jf\}_{j=0}^J\cup \{A_Jf\},$ denote a collection of wavelets constructed from a semigroup of operators $\{H^t\}_{t\geq 0}$. For $0\leq j \leq j'\leq J$ and $1\leq q\leq Q,$ we will let $Sf[q],$ $Sf[j,q],$ and $Sf[j,j',q]$ denote zeroth-, first-, and second-order scattering moments.

\section{The Manifold Scattering Transform}\label{sec: MSdef}

In this section, we present the \emph{manifold scattering transform}. Our construction is based upon the one presented in  \cite{perlmutter:geoScatCompactManifold2020} but with several modifications to increase computational efficiency and improve the expressive power of the network.  Namely, i) we use different wavelets, which can be implemented without the need to diagonalize a matrix (as discussed in Section \ref{sec: noeigenapprox}) and ii) we include $q$-th order scattering moments to increase the discriminative power of our representation.

Let $\mathcal{M}$ denote a smooth, compact,  connected $d$-dimensional Riemannian manifold without boundary isometrically embedded in $\mathbb{R}^n$. Since $\Mc$ is compact and connected, 
the (negative) Laplace Beltrami operator
$-\Delta$ has countably many eigenvalues, $0 = \lambda_0 < \lambda_1 \leq \lambda_2\leq \ldots$ and there exists a sequence of eigenfunctions such that $-\Delta \varphi_k=\lambda_k\varphi_i$ and  $\{ \varphi_k \}_{k \geq 0}$ forms an orthonormal basis for $\Lb^2 (\Mc)$.
Following the lead of works such as  \citet{shuman:emerging2013}, we will interpret the  eigenfunctions $\{ \varphi_k \}_{k \geq 0}$ as generalized Fourier modes  and define the Fourier coefficients of  $f \in \Lb^2 (\Mc)$ by
\begin{equation*}
    \hf (k) \coloneqq \langle f, \varphi_k \rangle_{\mathbf{L}^2(\mathcal{M})} \coloneqq \int_{\Mc} f(y) \overline{\varphi_k (y)} \, dy \, .
\end{equation*}
Since $\varphi_0,\varphi_1,\varphi_2,\ldots$ form an orthonormal basis, we have \begin{equation*}     f(x) =   \sum_{k \geq 0} \hf (k) \varphi_k(x). 
\end{equation*}

 In Euclidean space, the convolution of a function $f$ against a filter $h$ is defined by integrating $f$ against translations of $h$. While translations are not well-defined on a general manifold, convolution can also be characterized as multiplication in the Fourier  domain, i.e.,  $\widehat{f \ast h}(\omega) = \hf (\omega) \hh (\omega)$.  
Therefore, for  $f, h \in \Lb^2 (\Mc),$ we define the convolution of $f$ and $h$ as
\begin{equation} \label{eqn: convolution on M}
    f \ast h (x) \coloneqq \sum_{k \geq 0} \hf (k) \hh (k) \varphi_k (x). 
\end{equation}

In order to use this notion of spectral convolution to construct a semigroup $\{H\}_{t\geq 0}$,
we let $g:[0,\infty)\rightarrow [0,\infty)$ be a nonnegative, decreasing function with $g(0)=1$ and $g(t)<1$ for all $t>0$, and let 
%
 $H^t$ be the operator corresponding to convolution against $\sum_{k\geq 0} g(\lambda_k)^t\varphi_k$, i.e.,
\begin{equation}\label{eqn: h}
H^tf =  \sum_{k\geq 0} g(\lambda_k)^t\widehat{f}(k)\varphi_k.
\end{equation}
By construction, $\{H^t\}_{t\geq 0}$ forms a semigroup since $H^tH^s=H^{t+s}$ and $H^0=\text{Id}$ is the identity operator. 
We note that since the function $g$ is defined independently of $\mathcal{M}$, we may reasonably use our network, which is constructed from $\{H^t\}_{t\geq 0}$, to compare different manifolds. Moreover, it has been observed in \citet{levie2021transferability} that if
 $g$ is continuous, then $H^t$ is stable to small perturbations of the spectrum of $-\Delta$. 
We also note that one may imitate the proof of Theorem A.1 of \cite{perlmutter:geoScatCompactManifold2020} to verify that the definiton of $\{H_t\}_{t\geq 0}$ does not depend on the choice of eigenbasis $\{\varphi_k\}_{k\geq0}$.

As our primary example, we will take $g(\lambda)=e^{-\lambda}$, in which case,  one may verify that, 
for sufficiently regular functions,  $u_f(x,t)=H^tf(x)$ satisfies the heat equation
$$\partial_t u_f=-\Delta_xu,\quad  u(x,0)=f(x), $$
since we may compute 
\begin{align*}
    \partial_t H^tf(x) = \partial_t \sum_{k\geq 0} e^{-\lambda_kt} \widehat{f}(k)\varphi_k(x)
    =-\Delta_x H^tf(x).
\end{align*}
Thus, in this case $\{H^t\}_{t\geq 0}$ is known as the heat-semigroup. 


Given this semigroup, we define the wavelet transform  \begin{equation}\label{eqn: diffusion wavelets}
\mathcal{W}_Jf \coloneqq \{W_jf\}_{j=0}^J\cup \{A_Jf\},
\end{equation}
where $W_0\coloneqq\text{Id}-H^1,$ $A_{J}\coloneqq H^{2^J}$, and for $1\leq j \leq J$
\begin{equation}\label{eqn: Wj}
W_j \coloneqq H^{2^{j-1}}-H^{2^{j}}.
\end{equation}
These wavelets aim to track changes in the geometry of $\mathcal{X}$  across different diffusion time scales. Our construction uses a minimal time scale of $1$. However, if one wishes to obtain wavelets which are sensitive to smaller time scales, they may simply change the spectral function $g$. For example, if we set $g(\lambda)=e^{-\lambda}$ and $\tilde{g}(\lambda)=e^{-\lambda/2}$, then the associated operators satisfy $\widetilde{H}^1 = H^{1/2}$.

We note that this wavelets differ slightly from those considered in  \citet{perlmutter:geoScatCompactManifold2020}. For $1\leq j \leq J$, the wavelets $W_j'$ constructed there are given by 
 \begin{equation}\label{eqn: different wavelets}W'_jf(x) = \sum_{k\geq 0} \sqrt{g(\lambda_k)^{2^{j-1}}-g(\lambda_k)^{2^{j}}}\widehat{f}(k)\varphi_k.
\end{equation}
By contrast, the wavelets given by \eqref{eqn: Wj}
satisfy
$$ W_jf(x) = \sum_{k\geq 0} \left(g(\lambda_k)^{2^{j-1}}-g(\lambda_k)^{2^{j}}\right)\widehat{f}(k)\varphi_k.
$$
(Similar relations hold for $A_J$ and $W_0$.) The removal of the square root is to allow for efficient numerical implementation. In particular, in Section \ref{sec: noeigenapprox}, we will show that $H^1$ can be approximated without computing any eigenvalues or eigenvectors. Therefore, the entire wavelet transform can be computed without the need to diagonalize a (possibly large) matrix. We note that our wavelets, unlike those presented in \citet{perlmutter:geoScatCompactManifold2020},  are not an isometry. However, one may imitate the proof of Proposition 4.1 of \citet{gama:diffScatGraphs2018} or Proposition 2.2 of \citet{perlmutter2019understanding} to verify that our wavelets are a nonexpansive frame.

The scattering transform consists of an alternating sequence of wavelet transforms and nonlinear activations. As in most papers concerning the scattering transform, we will take our nonlinearity to be the pointwise absolute value operator $|\cdot|$. However, our method can easily be adapted to include other activation functions such as sigmoid, ReLU, or leaky ReLU.
 
For $0\leq j\leq J$ and $1\leq q\leq Q$, we define first-order $q$-th scattering moments by 
\begin{equation*}
    Sf[j,q]\coloneqq \int_\mathcal{M}|W_jf(x)|^qdx=\|W_jf\|_{\mathbf{L}^q(\mathcal{M})}^q,
\end{equation*}
and define second-order moments, for $0\leq j<j'\leq J$, by \begin{equation*}
    Sf[j,j',q]\coloneqq \int_\mathcal{M}|W_{j'}|W_jf(x)||^qdx=\|W_{j'}|W_jf|\|_{\mathbf{L}^q(\mathcal{M})}^q.
\end{equation*} 
Zeroth-order moments are defined simply by $$Sf[q]\coloneqq\int_\mathcal{M}|f(x)|^qdx=\|f\|_{\mathbf{L}^q(\mathcal{M})}^q.$$
We let $Sf$ denote the union of all zeroth-, first-, and second-order moments. If one is interested in classifying many signals, $\{f_i\}_{i=1}^{N_s}$, on a fixed manifold, one may compute $Sf_i$ for each signal and then feed these representations into a classifier such as a support vector machine. Alternatively, if one is interested in classifying different manifolds $\{\mathcal{M}_j\}_{j=1}^{N_M},$ one can pick a family of generic functions $\{f_i\}_{i=1}^{N_s}$ such as randomly chosen Dirac delta functions or SHOT descriptors \citep{tombari2010unique}. Concatenating the scattering moments of each $f_i$ gives a representation of each $\mathcal{M}_j$, which can be input to a classifier.
\section{Implementing the Manifold Scattering Transform from Pointclouds}\label{sec: implemention}
In this section, we will present methods for  implementing the manifold scattering transform on point cloud data. We will let $\{x_i\}_{i=1}^{N-1}\subseteq{\mathbb{R}^n}$ and assume that the $x_i$ are sampled from a $d$-dimensional manifold $\mathcal{M}$ for some $d\ll n$. As alluded to in the introduction, in the case, where $d=2,$ it is common (see e.g.  \citet{boscaini2015learning}) to approximate $\mathcal{M}$ with a triangular mesh. However, this method is not appropriate for larger values of $d$. Our methods, on the other hand, 
are valid for arbitrary values of $d$.

 Our first method relies on first computing an $N\times N$ matrix which approximates the Laplace Beltrami operator and then computing its eigendecomposition. It is motivated by results    which provides guarantees for the convergence of the eigenvectors and eigenvalues \citep{cheng2021eigen}. (See also \citet{DUNSON2021282}.) For large values of $N$, computing eigendecompositions is computationally expensive. Therefore we also present a second method which does not require the computation of any eigenvectors or eigenvalues. In either case, we identify $f$ with the vector $\mathbf{x}_f\in\mathbb{R}^N$, $\mathbf{x}_f(i)=f(x_i)$ and so once $H^t$ has been computed, it is then straightforward to implement the wavelet transform described in \eqref{eqn: diffusion wavelets} and \eqref{eqn: Wj} and then compute the scattering moments.

 We let $K:\mathcal{M}\times\mathcal{M}\rightarrow \mathbb{R}^{+}\cup\{0\}$ be an affinity kernel and given $K$, we define an affinity matrix $W$ and a diagonal degree matrix $D$ by
 \begin{equation}\label{eqn: DW}
     W_{i,j} = K(x_i,x_j) ,\quad\text{and}\quad D_{i,i}=\sum_{j=0}^{N-1}W_{i,j}.
 \end{equation}
\subsection{Approximating $H^t$ via the Spectrum of the Data-Driven Laplacian}\label{sec: eigenapprox}
In this section, for $\epsilon>0$, we define $K=K_\epsilon$ by 
\begin{equation}\label{eqn: guassian K} 
     K_\epsilon(x,x')=\epsilon^{-d/2}\exp\left(-\frac{\|x-x'\|^2_{2}}{\epsilon}\right).
 \end{equation}
 We then construct a discrete approximation 
 $-\Delta$ by 
$$ L^{N,\epsilon}=\frac{1}{\epsilon N} (D-W),$$
where $D$ and $W$ are as in \eqref{eqn: DW}.
We denote the eigenvectors and eigenvalues of $L^{N,\epsilon}$ by $\lambda_k^{N,\epsilon}$ and $\mathbf{u}_k^{N,\epsilon}$ so that
$
L^{N,\epsilon} \mathbf{u}_k^{N,\epsilon} = \lambda_k^{N,\epsilon} \mathbf{u}_k^{N,\epsilon}.
$ 
Using the first $\kappa$ eigenvectors and eigenvalues of $L^{N,\epsilon}$, 
 we define  
\begin{equation*}
    H^t_{N,\kappa,\epsilon} \coloneqq \sum_{k=0}^{\kappa-1} e^{-{\lambda_k^{N,\epsilon}t}} \mathbf{u}_k^{N,\epsilon}(\mathbf{u}_k^{N,\epsilon})^T.
\end{equation*}
To motivate this method, we note
Theorem 5.4 of  \cite{cheng2021eigen}, which shows that if we set
$\epsilon \sim N^{-1/(d/2+3)}$, assume that data points $\{x_i\}_{i=0}^{N-1}$ are sampled i.i.d. uniformly from $\mathcal{M}$ and certain other mild conditions hold, then
there exist scalars $\alpha_k$ with $|\alpha_k|=1+o(1)$ such that 
\begin{align}
    |\mu_k-\lambda^{N,\epsilon}_k|&=\mathcal{O}\left(N^{-\frac{1}{d/2+3}}\right),\label{eqn: eval rate}\\ \|\mathbf{u}_k^{N,\epsilon}-\alpha_k\mathbf{v}_k\|_2&=\mathcal{O}\left(N^{-\frac{1}{d/2+3}}\sqrt{\log N}\right),\label{eqn: evec rate}
\end{align}for $0\leq k\leq \kappa-1$
with probability at least $1-\mathcal{O}\left(\frac{1}{N^9}\right)$, where $\mu_k$ are the true eigenvalues of the Laplace Beltrami operator $-\Delta$, the $\mathbf{v}_k$ are vectors obtained by subsampling (and renormalizing) its eigenfuctions $\varphi_k$, and 
the implicit constants depend on both $\mathcal{M}$ and $\kappa$.

\subsection{Approximating $H$ without Eigenvectors}\label{sec: noeigenapprox}
The primary drawbacks of the method of the previous section are that it requires one to compute the eigendecomposition of $L^{N,\epsilon}$, and it  does not account for the  possibility that the data is sampled non-uniformly. 
One method for handling this problem is to use an adaptive method which scales the bandwidth of the kernel at each data point \citep{zelnik2004self,Cheng2021convergence}  
Here, we let $\sigma_k(x)$ denote the distance from $x$ to its $k$-th nearest neighbor and define an  adaptive kernel by
\begin{equation}\label{HS_kernel}
\begin{split}
&K_{k-nn}(x, x')=\\
&\frac{1}{2}\left(\exp\left(-\frac{\|x-x'\|^2_{2}}{\sigma_{k}(x)^2}\right) + \exp\left(-\frac{\|x-x'\|^2_{2}}{\sigma_{k}(x')^2}\right)\right)
\end{split}
\end{equation}

Then, letting $W$ and $D$ be as in \eqref{eqn: DW}, with $K=K_{k-nn}$, we define an approximation of $H^1$ by 
\begin{equation}\label{eqn: HapproxNoeigs}
    H^1_{N,\epsilon} \coloneqq D^{-1}W.
\end{equation}
We may then approximate $H^{2^j}$ via matrix multiplication without computing any eigenvectors or values. We note that while in principle $H^1_{N,\epsilon}$ is a dense matrix, by construction, most of its entries will be small (if $k$ is sufficiently small). Therefore, if one desires, they may use a thresholding operator to approximate $H^1_{N,\epsilon}$ by a sparse matrix for increase computational efficiency. 

The approximation \eqref{eqn: HapproxNoeigs} can be obtained by approximating $-\Delta$ with the Markov normalized graph Laplacian $I_N-D^{-1}W$ and then setting $g(\lambda)=1-\lambda$ in \eqref{eqn: h}.
We note that this  is similar to the approximation $H^1\approx D^{-1/2}WD^{-1/2}$ proposed in  \citet{coifman:diffusionMaps2006}, but  we use Markov normalization rather than symmetric normalization.  

\section{Related work}\label{sec: related}
In this section, we summarize related work on geometric scattering as well as neural networks defined on manifolds.

\subsection{Graph and Manifold Scattering}
Several different works \citep{zou:graphCNNScat2018,gama:diffScatGraphs2018,gao:graphScat2018} have introduced different versions of the graph scattering transform using wavelets based on  \citet{hammond:graphWavelets2011} or  \citet{coifman:diffWavelets2006}.  \citet{zou:graphCNNScat2018} and  \citet{gama:diffScatGraphs2018} provided extensive analysis of their networks stability and invariance properties. These analyses were later generalized in  \citet{gama:stabilityGraphScat2019} and  \citet{perlmutter2019understanding}.  \citet{gao:graphScat2018}, on the other hand, focused on empirically demonstrating the effectiveness of their network for graph classification tasks.

Several subsequent works have proposed modifications to the graph scattering transform for improved performance.  \citet{tong2020data} and  \citet{ioannidis2020pruned} have proposed data-driven methods for selecting the scales used in the network.  \citet{wenkel2022overcoming} incorporated the graph scattering transform incorporated into a hybrid network which also featured low-pass filters similar to those used in networks such as  \cite{kipf2016semi}.   We also note works  using the scattering transform for graph generation  \citep{zou:graphScatGAN2019,bhaskar2021molecular}, for solving combinatorial optimization problems \cite{Min2022MC}, and work  which extends the scattering transform to spatio-temporal graphs \citep{pan2020spatio}.

The work most closely related to ours is  \citet{perlmutter:geoScatCompactManifold2020} which introduced a different version of the manifold scattering transform. 
The focus of  this work was primarily theoretical, showing that the manifold scattering transform is invariant to the action of the isometry group and stable to diffeomorphisms, and the numerical experiments presented there are limited to two-dimensional surfaces with triangular meshes. Our work here builds upon \citet{perlmutter:geoScatCompactManifold2020} by introducing effective numerical methods for implementing the manifold scattering transform when the intrinsic dimension of the data is greater than two or when one only has access to the function on a point cloud. We also note  \citet{mcewen2021scattering}, which optimized the construction of  \citet{perlmutter:geoScatCompactManifold2020} for the sphere. 

\subsection{Manifold Neural Networks}
In contrast to graph neural networks, which are very widely studied (see, e.g.,  \citet{bruna:spectralNN2014,velivckovic2017graph,hamilton2017inductive}), there is comparatively little work developing neural networks for manifold-structured data. Moreover, unlike our method, much of the existing literature is either focused on a specific manifold of interest or is limited to two-dimensional surfaces.  

 \citet{Masci:geoCNN2015} defines convolution on two-dimensional surfaces via a method based on geodesics and local patches. They show that various spectral descriptors such as the heat kernel signature \citep{sun2009concise} and wave kernel signature \citep{aubry2011wave} can be obtained as special cases of their network. 
 In a different approach,  \cite{schonsheck2022parallel} defined convolution in terms of parallel transport. 
 \cite{boscaini2015learning}, on the other hand, used  spectral methods to define convolution on surfaces via a windowed Fourier transform. 
 This work was generalized in \citep{boscaini2016learning} to a network using an anisotropic Laplace operator. We note the many of the ideas introduced in    \citet{boscaini2015learning} and  \citet{boscaini2016learning} could in principle be applied to manifolds of arbitrary dimensions. However, their numerical methods  rely on triangular meshes. In more theoretical work,  \citet{wang2021stabilityrel,wang2021stability} also proposed networks constructed via functions of the Laplace Beltrami operator and analyzed the networks' stability properties.

In a much different approach,  \citet{chakraborty2020manifoldnet} proposed extending convolution to manifolds via a weighted Fr\'echet mean. Unlike  
the work discussed above,
this construction is not limited to two-dimensional surfaces. It does, however, require that the manifold is known in advance. We also note that there have been several papers which have constructed neural networks for specific manifolds such as the sphere \citep{cohen:sphericalCNNs2018}, Grassman manifolds  \cite{huang2018building}, or the manifold of positive semi-definite $n\times n$ matrices \citep{huang2017riemannian} or for homogeneous spaces \citep{chakraborty2018cnn,kondor:equivarianceNNGroups2018}.

\section{Results}\label{sec: results}
 We conduct experiments on  synthetic  and real-world data. These experiments aim to show that i) the manifold scattering transform is effective for learning on two-dimensional surfaces, even without a mesh and ii) our method is effective for high-dimensional biomedical data\footnote{All of our code is available \href{https://github.com/steachhr/pointcloud_scattering}{at our GitHub}.}.

\subsection{Two-dimensional surfaces without a mesh}

When implementing convolutional networks on a two-dimensional surfaces, it is standard, e.g., \citep{boscaini2015learning,boscaini2016learning} to use triangular meshes. 
In this section, we show that mesh-free methods can also work well in this setting. Importantly, we note that we are \emph{not} claiming that mesh-free methods are \emph{better} for two-dimensional surfaces. Instead, we aim to show that these methods can work relatively well thereby justifying their use in higher-dimensional 
settings. 

 We  conduct experiments using both mesh-based and mesh-free methods on a spherical version of MNIST and on the FAUST dataset which were previously considered in \citet{perlmutter:geoScatCompactManifold2020}. In both methods, we use the wavelets defined in Section \ref{sec: MSdef} with $J=8$ and use an RBF kernel SVM with cross-validated hyperparameters as our classifier. For the mesh-based methods, we use the same discretization scheme as in \citet{perlmutter:geoScatCompactManifold2020} and set $Q=1$ which was the setting implicitly assumed there. For our mesh-free experiments, we use the method discussed in Section \ref{sec: eigenapprox} with $Q=4$. We show that the information captured by the higher-order moments can help compensate for the structure lost by not using a mesh.

\begin{figure}
    \centering
    \includegraphics[width=0.450\textwidth]{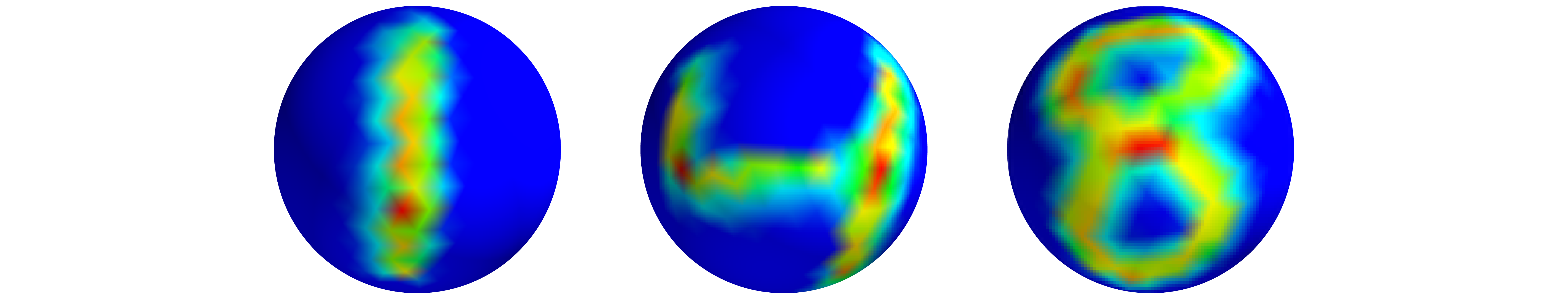}
    \caption{The MNIST dataset projected onto the sphere.}
    \label{fig:MNIST}
    \vskip -0.1in
\end{figure}


We first study the MNIST dataset projected onto the sphere as visualized in Figure \ref{fig:MNIST}. We uniformly sampled $N$ points from the unit two-dimensional sphere, and then applied random rotations to the MNIST dataset and projected each digit onto the spherical point cloud to generate a collection of signals $\{f_i\}$ on the sphere. Table \ref{mnist-table} shows that for properly chosen $\kappa$, the mesh-free method can achieve similar performance to the mesh-based method. As noted in Section \ref{sec: noeigenapprox}, the implied constants in \eqref{eqn: eval rate} and \eqref{eqn: evec rate} depend on $\kappa$ and inspecting the proof in \citet{cheng2021eigen} we see that for larger values of $\kappa,$ more sample points are needed to ensure the convergence of the first $\kappa$ eigenvectors. Thus, we want $\kappa$ to be large enough to get a good approximation of $H^1$, but also not too large. 
\begin{table}[t]
\caption{Classification accuracies for spherical MNIST.}
\label{mnist-table}
\vskip 0.15in
\begin{center}
\begin{small}
\begin{sc}
\begin{tabular}{ccccc}
\toprule
Data type & $N$ &$\kappa$ & $Q$ & Accuracy \\
\midrule
Point cloud    &1200 & 200& 4&  79\%\\
Point cloud    &1200 & 400& 4&  88\%\\
Point cloud    &1200 & 642& 4&  84\%\\
Mesh & 642 & 642 & 1 & 91\%\\

\bottomrule
\end{tabular}
\end{sc}
\end{small}
\end{center}
\vskip -0.1in
\end{table}

\begin{figure}
    \centering
    \includegraphics[width=0.450\textwidth]{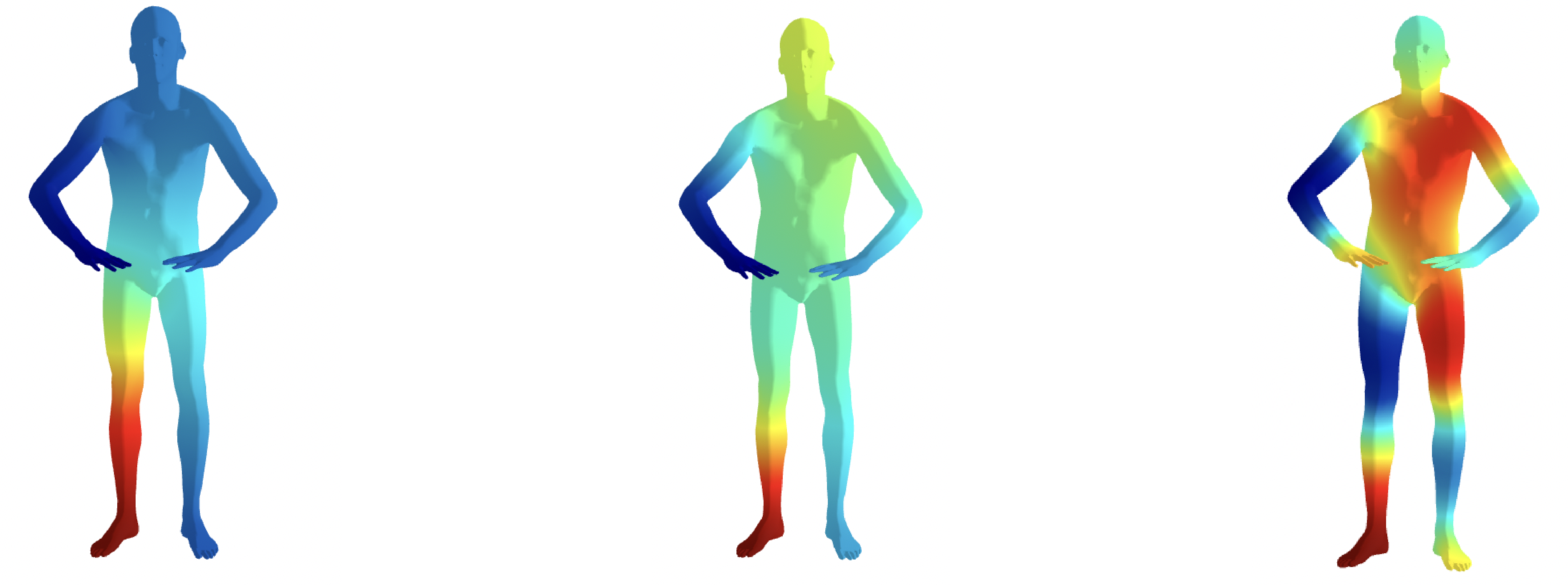}
    \caption{Wavelets on the FAUST dataset, $j = 1, 3, 5$ from left to right. Positive values are  red, while negative values are blue.}
    \label{fig:geometric wavelets faust}
\end{figure}

Next, we consider the FAUST dataset, a collection of surfaces corresponding to scans of ten people in ten different poses \citep{Bogo:CVPR:2014} as shown in Figure \ref{fig:geometric wavelets faust}. As in \citet{perlmutter:geoScatCompactManifold2020}, we use 352 SHOT descriptors \citep{tombari2010unique} as our signals. 
We used the first $\kappa = 80$ eigenvectors and eigenvalues of the approximate Laplace-Beltrami operator of each point cloud to generate scattering moments. We achieved 95\% classification accuracy for the task of classifying different poses. This  matches the accuracy obtained with meshes in \citet{perlmutter:geoScatCompactManifold2020}.

\subsection{Single-cell datasets}

In this section, we present two experiments leveraging the inherent manifold structure in single-cell data to show the utility of manifold scattering in describing and classifying biological data. In these experiments, single-cell protein measurements on immune cells obtained from either melanoma or SARS-COV2 (COVID) patients are featurized using the log-transformed L1-normalized protein expression feature data of the cell. 
This leads to one point in high-dimensional space corresponding to each cell, and 
we model the set of points corresponding to 
a given patient as lying on a manifold representing the immunological state of that patient. 
 In this context, it is not reasonable to assume that data are sampled uniformly from the underlying manifold $\mathcal{M}$ so we use the method presented in Section \ref{sec: noeigenapprox}  to approximate $H^t$.
 
 For both experiments, we set $k=3$,  use third-order scattering moments with $J=8$ and $Q=4$, reduce dimensions of scattering features with PCA, and use the top 10 principle components for classification with a decision tree. As a baseline comparison, we perform K-means clustering on all cells 
 and predict patient outcomes based on cluster proportions. 


We first consider data from \citet{PtacekA59}. In this dataset, 11,862 T lymphocytes from core tissue sections were taken from 54 patients diagnosed with various stages of melanoma and  30 proteins were measured per cell. Thus, our dataset consists of 54 manifolds, embedded in 30-dimensional space with 11,862 points per manifold. 
All patients received checkpoint blockade immunotherapy, which licenses patient T cells to kill tumor cells. Our goal is to characterize the immune status of patients prior to  treatment and to predict which patients will respond well. 
In our experiments, we first computed a representation of each patient via  either K-means cluster proportions, with $K=3$ based on expected T cell subsets (killer, helper, regulatory), then calculated scattering moments for protein expression feature signals projected onto the cell-cell graph. We achieved 46\% accuracy with clustering and 82\% accuracy with scattering. 

We next analyze 148 blood samples from COVID patients and focus on innate immune (myeloid) cells, a population that we have previously shown to be predictive of patient mortality \citep{RN3}. 14 proteins were measured on 1,502,334 total monocytes. To accommodate the size of these data, we first perform diffusion condensation \citep{RN3} for each patient dataset; we determine the lowest number of steps needed to condense data into less than 500 clusters and use those cluster centroids as single points in high-dimensional immune state space. Protein expression is averaged across cells in each cluster, and diffusion scattering is performed on these feature signals projected onto the centroid graph.  For K-means we use $K=3$ based on expected monocyte subtypes (classical, non-classical, intermediate). When used to predict patient mortality, we achieved 40\% accuracy with cluster proportions and 48\% accuracy with scattering features. 

\begin{table}[t]
\caption{Classification accuracies for patient outcome prediction.}
\label{melanoma-table}
\vskip 0.15in
\begin{center}
\begin{small}
\begin{sc}
\begin{tabular}{lcccr}
\toprule
Data set & $N_M$ & Clustering & Scattering \\
\midrule
Melanoma    &54 & 46\% & 82\% \\
COVID &148 &40\% & 48\% \\
\bottomrule
\end{tabular}
\end{sc}
\end{small}
\end{center}
\vskip -0.1in
\end{table}

\section{Conclusions}\label{sec: conclusion}
We have presented two efficient numerical methods for implementing the manifold scattering transform from point cloud data. Unlike most other methods proposing neural-network like architectures on manifolds, we do not need advanced knowledge of the manifold, do not need to assume that the manifold is two-dimensional, and do not require a predefined mesh. We have demonstrated the numerical effectiveness of our method for both signal classification and manifold classification tasks. 

\section*{Acknowledgements}
Deanna Needell and Joyce Chew were partially supported by NSF DMS 2011140 and NSF DMS 2108479. Joyce Chew was additionally supported by the NSF DGE 2034835. Matthew Hirn and Smita Krishnaswamy were partially supported by NIH grant R01GM135929. Matthew Hirn was additionally supported by NSF grant DMS 1845856.
 Smita Krishnaswamy was additionally supported by NSF Career grant 2047856 and Grant No. FG-2021-15883 from the Alfred P. Sloan Foundation. Holly Steach was supported by NIAID grant T32 AI155387. 
\bibliography{main}
\bibliographystyle{icml2021}

\end{document}